\documentclass{article}
\usepackage{ijcai21}

\usepackage{times}

\usepackage{soul}
\usepackage{url}
\usepackage[hidelinks]{hyperref}
\usepackage[utf8]{inputenc}
\usepackage[small]{caption}
\usepackage{graphicx}
\usepackage{amsmath}
\usepackage{booktabs}

\usepackage{times}
\usepackage{epsfig}
\usepackage{graphicx}
\usepackage{amsmath}
\usepackage{amssymb}
\usepackage{multirow}
\usepackage{color}
\usepackage{algorithmic}
\usepackage{algorithm}
\usepackage{booktabs}
\usepackage[table]{xcolor}
\usepackage{colortbl}
\usepackage{bm}
\usepackage{subcaption}
\usepackage{makecell}
\usepackage{pifont}% http://ctan.org/pkg/pifont

\usepackage{threeparttable}
\usepackage{stfloats}

\usepackage{enumitem}

\newcommand{\cmark}{\ding{51}}%
\newcommand{\xmark}{\ding{55}}%

\newcommand{\colorpm}[1]{\footnotesize{\textcolor[RGB]{113,117,250}{#1}}}

\title{Learning Implicit Temporal Alignment for Few-shot Video Classification} %[Action Recognition]
\author{
Songyang Zhang$^{1,2,4,}$\footnote{Equal contribution. This work was supported by Shanghai NSF Grant (No. 18ZR1425100). Code and model are available: \href{https://github.com/tonysy/PyAction}{https://github.com/tonysy/PyAction}}\and
Jiale Zhou$^{1,*}$\and
Xuming He$^{1,3}$\\
\affiliations
$^1$ShanghaiTech University\quad $^2$ University of Chinese Academy of Sciences\\ $^3$Shanghai Engineering Research Center of Intelligent Vision and Imaging\\$^4$Shanghai Institute of Microsystem and Information Technology, Chinese Academy of Sciences\\
\emails
\{zhangsy1,zhjl,hexm\}@shanghaitech.edu.cn
}

\begin{document}

\maketitle

\begin{abstract}
Few-shot video classification aims to learn new video categories with only a few labeled examples, alleviating the burden of costly annotation in real-world applications. However, it is particularly challenging to learn a class-invariant  spatial-temporal representation in such a setting. To address this, we propose a novel matching-based few-shot learning strategy for video sequences in this work. 
Our main idea is to introduce an implicit temporal alignment for a video pair, capable of estimating the similarity between them in an accurate and robust manner. Moreover, we design an effective context encoding module to incorporate spatial and feature channel context, resulting in better modeling of intra-class variations. 
To train our model, we develop a multi-task loss for learning video matching, leading to video features with better generalization. 
Extensive experimental results on two challenging benchmarks, show that our method outperforms the prior arts with a sizable margin on Something-Something-V2 and competitive results on Kinetics.
\end{abstract}

\section{Introduction}\label{sec:introduction}
% Video task is important and challenging
% Especially in few-shot settings
Video classification, which aims to learn spatio-temporal visual concepts in video, is a fundamental task in computer vision. In particular, recognizing action categories plays an important role in a wide range of applications such as video retrieval, behavior analysis and human-computer interaction. %~\cite{review}. 
Recently, significant progress has been made in video classification thanks to the powerful representation learned by deep neural networks~\cite{wang2016temporal,carreira2017quo,wang2018non}. 
While such fully-supervised approaches are capable of modeling class-specific short-term motion patterns and long-range temporal context, they typically require abundant annotated data. This is not only laborious due to extra temporal dimension, but also restrictive for deploying algorithms in real-world settings, which often need to adapt to novel categories.  
%(e.g., thousands of labeled videos)

% Formulate the problem as few-shot video action and introduce recent works
To tackle the issue of label scarcity, a prevailing strategy, inspired by human visual recognition, is to enable the algorithm to learn to recognize a novel video category with only a few annotated clips. Such a learning task, termed as \textit{few-shot video classification}, has attracted much attention recently~\cite{zhu2018compound,cao2020few,zhang2020few}. 
Most of these attempts adopt the metric-based meta-learning framework~\cite{vinyals2016matching}, in which they first match learned features from support and query video clips, and then assign the category label for query video based on the matching scores. 

Despite their promising results, those methods often suffer from two drawbacks due to the large intra-class variation of video categories and difficulty in matching two mis-aligned video instances. 
First, most approaches focus on learning a global video representation by exploiting salient memory mechanism~\cite{zhu2018compound} or introducing the spatial-temporal attention~\cite{zhang2020few}, which rarely explores the temporal alignment for matching videos. Second, while few recent method~\cite{cao2020few}  attempts to introduce explicit temporal alignment based on {dynamic time warping}~\cite{muller2007dynamic}, such matching process needs to calculate a dense frame-wise correspondence, which can be computationally expensive (e.g. in the square of video length).

In this work, we aim to tackle the aforementioned limitations by proposing a novel matching-based few-shot video classification strategy, which enables us to leverage the temporal structure of video instances and avoid dense frame correspondence estimation. Our main idea is to learn an implicit alignment of video sequences based on a factorized self-attention mechanism that enhances video representation along temporal, spatial and feature channel dimensions. Such augmented representation allows us to compute the similarity between two videos by directly summing similarity scores of their corresponding frames. More importantly, our attention-based feature augmentation and similarity metric are class-agnostic and can be learned via a meta-learning framework.   

Specifically, we represent each video as a fixed-length sequence of frame features, each encoded as a convolutional feature map. Our first step is to introduce a self-attention on each frame's feature map and a feature channel attention, which encode per-frame spatial context and a global video context, respectively. To achieve implicit temporal alignment, we further adopt a self-attention mechanism in temporal domain, which augments frame features with their temporal context and thus reduces temporal variation in video instances.    
We sequentially apply those three steps of attention-based feature augmentation, each of which can be computed efficiently. The final augmented video features are then used for computing a frame-based similarity in video matching.  
 
To perform few-shot learning, we develop a meta-learning strategy that employs a multi-task loss to simultaneously exploit the meta-level and semantic-level supervision. 
%This effectively regularizes the feature learning and speedup the overall meta-training process.
This effectively regularizes the feature learning during the meta-training process.
We validate our method on two video classification benchmarks, including Kinetics~\cite{Carreira_2017_CVPR} and Something-Something-V2~\cite{goyal2017something}. We also report an empirical study of different meta-learning strategies with varying loss designs.   

The main contributions of this work are summarized as the following:
\begin{itemize}[itemsep=0.25mm,topsep=0.5mm]
    \item We propose a simple and yet effective attention-based representation for few-shot video classification, which leverages temporal structure of videos and enables fast estimation of video similarity. 
    \item Our factorized self-attention mechanism captures the spatial, feature channel and temporal context of a video, leading to an implicit alignment effect in matching.   
    %\item We introduce a simple and effective context-aware video embedding by modeling the spatial context and feature channel context jointly.
%    \item A multi-task learning strategy are adopted for few-shot video classification, which improves training efficiency and model generalization.
    \item We adopt a multi-task learning strategy for few-shot video classification that improves model generalization.
    \item Our model achieves the state-of-the-art or competitive performance on two challenging benchmarks, Something-Something-V2 and Kinetics.
%    \item , and outperforms previous methods by a large margin. 
\end{itemize}
\section{Related Work}\label{sec:related}
\paragraph{Few-shot Learning}
Inspired by data-efficient learning in human cognition, few-shot learning aims to learn a new concept representation from only a few training examples. Most of existing works can be categorized into optimization-learning based~\cite{finn2017model,nichol2018reptile,ravi2016optimization,munkhdalai2017meta}, metric-learning based~\cite{vinyals2016matching,snell2017prototypical,sung2018learning} and graph neural network based methods~\cite{garcia2017few}.
%Such a learning paradigm has attracted much attention in the literature

% Optimization Based: Initialization Based & Optimizer Based
%For optimization-learning based methods, one typical branch, represented by MAML~\cite{finn2017model}, Reptile~\cite{nichol2018reptile}, aims at learning model initialization that achieves better performance after a few number of gradient update steps.
%~\cite{ravi2016optimization} proposes an LSTM-based meta-learner to learn the exact optimization algorithm. Meta Networks~\cite{munkhdalai2017meta} learns meta-level knowledge across tasks and shifts its inductive biases via fast parameterization for rapid generalization.

% Metric Learning Based
%Another type of approach, known as metric learning, focuses on learning a proper distance function that generalizes well on novel classes. 
Our work is inspired by the metric-learning based methods. In particularly, Matching Networks~\cite{vinyals2016matching} and Prototypical Networks~\cite{snell2017prototypical,liu2020part} learn discriminative feature representations of classes under cosine or Euclidean distance metric. Relation Network~\cite{sung2018learning} proposes a relation module as a learnable metric.
% Hallucination Based
%An alternative way is to directly deal with data deficiency by leveraging augmented data~\cite{hariharan2017low,wang2018low} or utlizing the task context with graph neural network~\cite{garcia2017few}
%In this work, we follow the metric learning manner to perform few-shot video classification.
Our work is in line with the Matching Networks, but we adopt this idea in more challenging video recognition task, focusing on a simple and effective design for measuring video similarity.

\paragraph{Video Action Recognition}
Video classification methods have evolved from using
hand-crafted features~\cite{klaser2008spatio,scovanner20073,wang2013action} to learning representations via deep networks. Particularly,
C3D~\cite{tran2015learning} utilizes 3D spatio-temporal convolutional filters to extract deep features from sequences of RGB frames. TSN~\cite{wang2016temporal} and I3D ~\cite{carreira2017quo} uses two-stream 2D or 3D CNNs on both RGB and optical flow sequences. 
Prior works~\cite{wang2016temporal,tran2015learning} on deep learning based video classification focus on modeling short-term action-specific motion information and long-range temporal context\cite{wang2018non,wang2016temporal,carreira2017quo}, which typically require abundant annotated data to train their models. Hence it is non-trivial to apply them in the few-shot learning setting directly. Moreover, our work is also related to some recent works aiming to reduce the computation cost of the non-local or self-attention\cite{chen2019graph,zhang2019latentgnn}.

%An issue of these video representation learning methods is their dependence on large-scale video datasets for training. Models with an excessive amount of learnable parameters tend to fail when only a small number of training samples are available. 

%Recently, non-local neural networks ~\cite{wang2018non} introduce self-attention to aggregate temporal information in the long-term. Wang et al. ~\cite{wang2018videos} further employ space-time region graphs to perform spatiotemporal reasoning. Recently, TRN ~\cite{zhou2018temporal} proposes a temporal relational module to achieve superior performance.

\paragraph{Few-shot Video Action Classification}
% cmn tarn otam 
%% TODO %%: add other RECENT works on video fsl
Few-shot video classification aims to classify new video classes with only a few annotated examples, and attracted much attention in community recently. The previous approaches mainly focus on the metric-learning based methods.
OSS-Metric Learning~\cite{kliper2011one} measures OSS-Metric of video pairs to enable one-shot video classification. \cite{mishra2018generative} learns a mapping function from an attribute to a class center for zero/few-shot video learning. CMN~\cite{zhu2018compound} improves the feature representation by introducing a multi-saliency embedding algorithm to encode video frames into a fixed-size matrix. \cite{zhang2020few} propose a model that captures short-range dependencies with a 3D backbone and discards long-range dependencies via a self-trained permutation invariant attention. These works typically use the global averaged feature for the final matching, which leave the temporal alignment rarely explored.
%They then propose a compound memory network (CMN) to store the representation and classify videos by matching and ranking.

More recent works have benefited from treating the video matching problem as the sequence matching problem. Inspired from text matching in machine translation~\cite{bahdanau2014neural,wang2016compare}, TARN~\cite{bishay2019tarn} utilizes attention mechanisms to perform temporal alignment. OTAM~\cite{cao2020few} suggests preserving the temporal order of video data and proposes an ordered temporal alignment module, which utilizes a differentiable variant of the classical DTW~\cite{muller2007dynamic} algorithm, to perform the matching. However, this kind of explicit alignment design needs to calculate a dense frame-wise similarity, making it inefficient in the video classification tasks.
%(e.g. in size of $T\times T$).
%All these matching-based ideas are limited by their heuristic design for the temporal alignment. 
By contrast, we propose a learnable implicit temporal alignment strategy to achieve accurate and robust video matching.
%n this wor

%\paragraph{Attention Mechanism}
%Our work is related to the recent self-attention ~\cite{vaswani2017attention} mechanism for machine translation. A self-attention module computes the response at a position in a sequence (e.g., a sentence) by attending to all positions and taking their weighted average in an embedding space.
%% TODO: variation of attention modules

\section{Problem Setup and Terminology}\label{sec:setting}
We consider the problem of few-shot video classification, which aims to learn to predict video categories from only a few annotated training video clips per category. To this end, we adopt a mete-learning strategy~\cite{vinyals2016matching} to build a meta learner $\mathcal{M}$ that solves similar few-shot video recognition tasks $\mathcal{T}=\{T\}$ sampled from an underling task distribution $P_T$. 

Formally, each few-shot video classification task $T$ (\textit{also called an episode}) is composed of a set of support data $\mathcal{S}$ with ground-truth labels and a set of query video clips $Q$. For the $C$-way $K$-shot setting, the annotated support data consists of $K$ clip-label pairs from each class, denoted as $\mathcal{S}=\{(\mathbf{X}^s_{c,k}, y^s_{c,k})\}_{k=1,\cdots,K}^{c\in \mathcal{C}_T}$, where $\{y_{c,k}\}$ are video-wise labels. $\mathcal{C}_T$ is the subset of class sets for the task $T$ and $|\mathcal{C}_T|=C$. Similarly, the query set $Q=\{(\mathbf{X}_{j}^q,y_j^q)\}_{j=1}^{N_q}$, contains $N_q$ videos from the class set $\mathcal{C}_T$ whose ground-truth labels $\{y_j^q\}$ are provided during training but \textit{unknown} in test.

The meta learner $\mathcal{M}$ aims to learn a functional mapping from the support set $\mathcal{S}$ and a query video $\mathbf{X}^q$ to its category $y^q$ for all the tasks. To achieve this, we construct a training set of video recognition tasks $\mathcal{D}^{tr}=\{(\mathcal{S}_n,\mathcal{Q}_n)\}_{n=1}^{|\mathcal{D}^{tr}|}$ with a class set $\mathcal{C}^{tr}$, and train the meta learner episodically on the tasks in $\mathcal{D}^{tr}$. After the meta-training, the model $M$ encodes the knowledge on how to perform video classification on different video categories across tasks. We finally evaluate the learned model on a test set of tasks $\mathcal{D}^{te}=\{(\mathcal{S}_m, \mathcal{Q}_m)\}_{m=1}^{|\mathcal{D}^{te}|}$ whose class set $\mathcal{C}^{te}$ is non-overlapped with $\mathcal{C}^{tr}$.

\begin{figure*}
    \centering
    \includegraphics[width=0.85\linewidth]{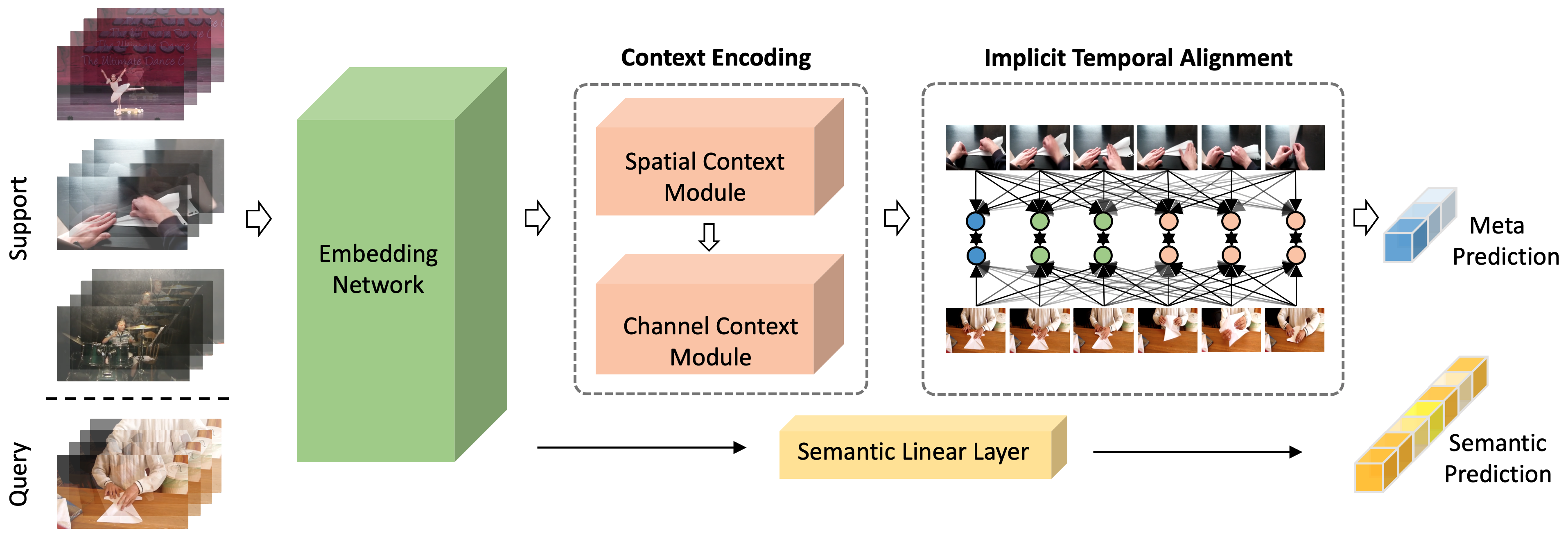}
%    \vspace{-2em}
    \caption{\textbf{Illustration of the overall framework.} In training phase, the semantic prediction and meta prediction are used for multi-task learning simultaneously. The semantic prediction is ignored during the inference on the given task. }\vspace{-3mm}
    \label{fig:overview}
\end{figure*}

\section{Our Approach}\label{sec:method}
In this work, we adopt a matching-based few-shot learning framework to build a meta-learner $\mathcal{M}$ for video classification. Our goal is to design a class-agnostic matching strategy capable of measuring the similarity of two video instances in an efficient and robust manner. To achieve this, we develop a video representation based on self-attention mechanism, which provides an implicit temporal alignment effect and yields an efficient matching-based video classifier. 

Specifically, we introduce a factorized self-attention mechanism that augment the original frame-based video representation in temporal, spatial and feature channel dimensions. This alleviates the intra-class variations of video instances, particularly along the temporal dimension. Consequently, we are able to compute the similarity of two videos via a direct summation of similarity scores of their corresponding frame-level features and thus bypass explicit video alignment. Such feature augmentation strategy does not depend on specific video class and can be generalized to novel categories.   

%The main idea of our method is to measure the two video clips accurately and robustly by introducing an implicit temporal alignment. Specifically, we propose to project the all frames of one clip into several latent frames, which enables the support video and query video are frame-wise aligned. 

To this end, we develop a deep conv network, as our meta-learner, to embed each video into its representation and to classify query videos. 
Our network consists of three main modules: an \textit{embedding network} that computes convolutional feature maps for the video frames within a task (in Sec.~\ref{sec:embedding}); a \textit{context encoding} network that enrich the feature representation of support and query videos by encoding the spatial and feature channel context (in Sec.~\ref{sec:context}); and an \textit{implicit temporal alignment network} to align the two videos for accurate and robust matching (in Sec.~\ref{sec:alignment}).

%To train our meta model, we introduce a multi-task learning strategy for the few-shot video recognition that exploits the entire set of known categories for efficient learning (in Sec.~\ref{sec:learning}).
To train our meta model, we introduce a multi-task learning strategy for the few-shot video recognition that exploits the entire set of known categories for effective learning (in Sec.~\ref{sec:learning}).
We refer to our model as the \textbf{Implicit Temporal Alignment Network (ITANet)}. An overview of our framework is illustrated in Fig.~\ref{fig:overview} and we will present the model details in the remaining of this section.

\subsection{Embedding Network}\label{sec:embedding}
Given a few-shot task (or episode), the first module of our ITANet is an embedding network that extracts the convolutional feature maps of all frames in each video clip. Following prior work~\cite{cao2020few,zhu2018compound}, we adopt ResNet~\cite{he2016deep} as our embedding network. Formally, we denote the embedding network as $f_{em}$, and compute the feature maps of videos in a task $T$ as $\mathbf{F}=f_{em}(\mathbf{X}), \forall \mathbf{X}\in \mathcal{S}\cup\mathcal{Q}$. Here $\mathbf{F}\in\mathbb{R}^{H_f\times W_f\times n_t\times n_{c}}$, $n_{c}$ is the number of feature channels, $(H_f,W_f,n_t)$ is the height, width and temporal length of the feature map.

\subsection{Context Encoding Network}\label{sec:context}
To cope with intra-class variation of videos, we first introduce a context encoding network to augment the video features with spatial context in each frame and global context of the entire video. Such a context-aware representation enriches the frame features and focuses more on the discriminative part of the video via attention mechanism. This is particularly important for the few-shot task. Specifically, the context encoding network consists of two components: a \textit{frame-wise spatial context module} and a \textit{video-wise channel context module}, which are described in detail as following:

\paragraph{Frame-wise Spatial Context Module} The frame-wise spatial context module augments the feature representation of each frame with its spatial context via a multi-head self-attention. Specifically, we have the feature map of one video clip as $\mathbf{F}=\{\mathbf{f}_{1},\cdots,\mathbf{f}_{n_t}\},\mathbf{f}_{n_t}\in \mathbb{R}^{N_f\times n_{c}}, N_f=H_f\times W_f$. For notation clarity, we focus on a single frame $\mathbf{f}$ here.

We adopt the self-attention mechanism as in~\cite{vaswani2017attention}, which maps queries and a set of key-value pairs to outputs. Formally, given the query $\mathbf{q}$, key $\mathbf{k}$ and value $\mathbf{v}$, we compute the attention as follows:
\begin{align}
	\text{Attention}(\mathbf{q,k,v})&=\text{softmax}(\frac{\mathbf{q}\mathbf{k}^\intercal}{\sqrt{d_k}})\mathbf{v}
\end{align}
Specifically, we employ the multi-head attention, which linearly projects the frame feature $\mathbf{f}$ by $h$ times with multiple learnable linear transforms to form queries, keys and values of $d_q,d_k$ and $d_v$ dimensions. The output of multi-head attention is generated by concatenating $h$ heads, followed by a linear transform:
\begin{align}
	\text{Head}_i& = \text{Attention}(\mathbf{f}\mathbf{W}_i^q,\mathbf{f}\mathbf{W}_i^k,\mathbf{f}\mathbf{W}_i^v),_{\;1\leq i\leq h}\label{eq:head_i}\\	\text{MultiHead}(\mathbf{f})&=\text{Concat}(\text{Head}_1,\cdots,\text{Head}_h)\mathbf{W}^o \label{eq:multihead}
\end{align}
where $\mathbf{W}_i^q\in\mathbb{R}^{n_{c}\times d_q}$, $\mathbf{W}_i^k\in\mathbb{R}^{n_{c}\times d_k}$, $\mathbf{W}_i^v\in\mathbb{R}^{n_{c}\times d_v}$ are the parameters of linear projections, $i$ is the index of the head,
and $\mathbf{W}^o\in\mathbb{R}^{hd_v\times n_{c}}$, $n_{c}$ is the feature dimension of the final outputs. 
We typically set $d_q=d_k=d_v=n_{c}/h$ and use the residual connection to generate the final spatial context-aware representation as:
\begin{align}
	\mathbf{f}^{sp}=\text{MultiHead}(\mathbf{f}) + \mathbf{f},\quad
	\mathbf{F}^{sp}=\{\mathbf{f}^{sp}_{1},\cdots,\mathbf{f}^{sp}_{n_t}\}
%	,\mathbf{f}_{n_t}\in \mathbb{R}^{N_f\times n_{c}}, N_f=H_f\times W_f
\end{align}
%\rev{Define MultiHead for the entire video here ???}.

\begin{figure}[t]
    \centering
    \includegraphics[width=0.9\linewidth]{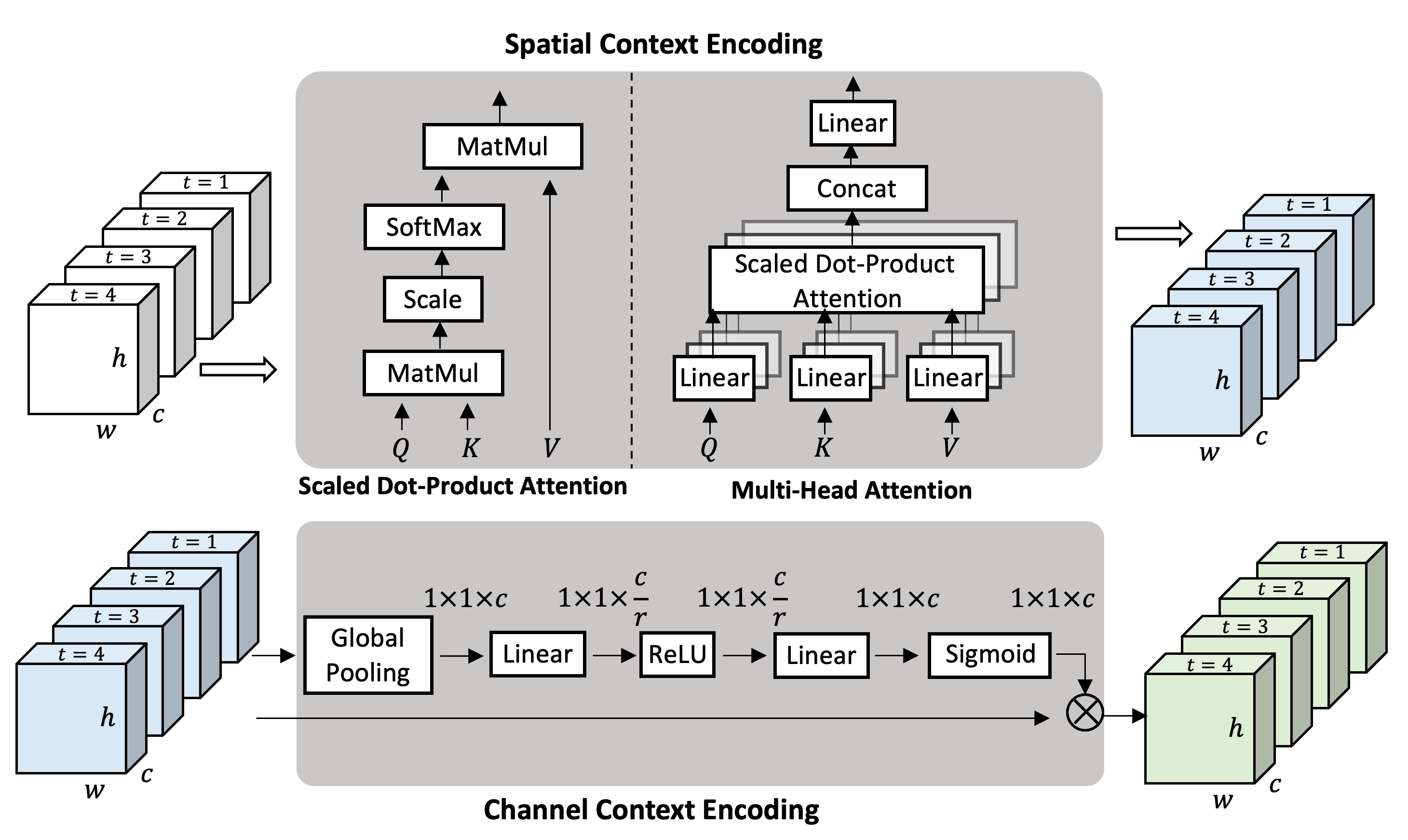}
%    \vspace{-2em}
    \caption{\textbf{Illustration of the Context Encoding Module.}}\vspace{-3mm}
    \label{fig:context_encoding}
\end{figure}

\paragraph{Video-wise Channel Context Module}
Inspired by the SENet~\cite{hu2018squeeze}, we develop a video-wise channel context module, which captures the interdependencies between the feature channels of a whole video, as illustrated in Fig.~\ref{fig:context_encoding}.
Formally, given feature representation of one video clip $\mathbf{F}^{sp}\in\mathbb{R}^{H_f\times W_f\times n_t\times n_{c}}$, we first squeeze the global spatial-temporal information into a video-wise channel descriptor as:
\begin{align}
	\mathbf{z} = \frac{1}{H_f\times W_f\times n_t}\sum_{i=1}^{H_f}\sum_{j=1}^{W_f}\sum_{k=1}^{n_t}\mathbf{F}_{(i,j,k)}\in\mathbb{R}^{n_{c}}
\end{align}
Then an adaptive recalibration is applied with excitation operation via a bottleneck structure as following:
\begin{align}
\mathbf{s}^{ch} = \sigma(\mathbf{W}_2^{ch}\text{ReLU}(\mathbf{W}_1^{ch}\mathbf{z}))\in\mathbb{R}^{n_c}
\end{align}
where $\mathbf{W}^{ch}_2\in \mathbb{R}^{n_{c}\times\frac{n_{c}}{r}}$ and $\mathbf{W}^{ch}_1\in \mathbb{R}^{\frac{n_{c}}{r}\times n_{c}}$ are fully-connected layers, $r$ is the dimensionality reduction ratio. $\text{ReLU}$ and $\sigma$ are the ReLU function and sigmoid function, respectively.
With the activation $\mathbf{s}^{ch}$, the channel context-aware representation are generated by element-wise multiplication along the channel dimension:
\begin{align}
	\widetilde{\mathbf{F}}=\{\mathbf{F}^{sp}_{1}\mathbf{s}^{ch}_1,\cdots,\mathbf{F}^{sp}_{n_{c}}\mathbf{s}_{n_{c}}^{ch}\}\in\mathbb{R}^{H_f\times W_f\times n_t\times n_c}
\end{align}
where $\mathbf{F}_{n_{ch}}^{sp}\in\mathbb{R}^{H_f\times W_f\times n_t}$ is the feature map of a single channel for the entire video.

\subsection{Implicit Temporal Alignment Network}\label{sec:alignment}
After integrating the spatial-channel context, we introduce another feature augmentation along temporal dimension to reduce temporal variation in videos, which enables us to achieve an implicit alignment effect on two videos for similarity estimation. To this end, we design an implicit temporal alignment network for aligning and matching a pair videos, consisting of two modules as follows: a) a \textit{temporal relation encoding module} to enrich each frame with temporal relational context; b) a \textit{similarity metric module} to compute the similarity of a query and a support video.

\paragraph{Temporal Relation Encoding Module} We first squeeze the spatial dimension of context encoded feature map $\widetilde{\mathbf{F}}$ with average pooling operation to generate frame-level representation as follows:
\begin{align}
	\mathbf{Z}&= \text{AvgPool}_{spatial}(\widetilde{\mathbf{F}})\\
	\mathbf{Z}&=\{\mathbf{z}_1,\cdots,\mathbf{z}_{n_t}\}\in\mathbb{R}^{n_t\times n_{c}}
\end{align}
%Specifically, we first define a set of the frames vectors $\mathbf{U}=\{\mathbf{u}_1,\cdots,\mathbf{u}_d\}\in \mathbb{R}^{d\times n_{c}}$ for one clip as the latent frames. 
To encode temporal order information, we also add position encoding~\cite{vaswani2017attention} to update the frame-level feature $\mathbf{Z}$ as below:
\begin{align}
	\mathbf{z}_{n_t}^p&=\mathbf{z}_{n_t} + \mathbf{p}_{n_t},\quad \mathbf{p}=f_{pos}(t)\in\mathbb{R}^{n_c}\\
	\mathbf{Z}^{p} &=\{\mathbf{z}_1^p,\cdots,\mathbf{z}_{n_t}^p\}\in\mathbb{R}^{n_t\times n_{c}}
\end{align}
where $f_{pos}$ is position encoding function, which can be cosine/sine function or learnable function.
%a learnable non-linear transformation to embed the temporal order into a feature space.

Given the video representation $\mathbf{Z}^p\in\mathbb{R}^{n_t\times n_{c}}$, we apply the multi-head self-attention, in a similar form detailed in Eq~\ref{eq:head_i} and Eq.~\ref{eq:multihead}, to all the frame of a video to enhance each frame with temporal relational context. We also use the residual structure to produce the final video representation as follows:
\begin{align}
	\widetilde{\mathbf{Z}}=\text{MultiHead}(\mathbf{Z}^p) + \mathbf{Z}^p
\end{align}
\paragraph{Similarity Metric Module}
Given our enriched video representation, we measure the similarity of two video clips by direct summing the similarity scores of their corresponding frame features. Such a similarity metric also facilitates learning an implicit temporal alignment for video matching, which is enabled by our multi-dimension feature augmentation. 

Concretely,  given two videos $\mathbf{X}_i$, $\mathbf{X}_j$, their augmented representations are denoted as $\mathbf{\tilde{Z}}^i=\{\mathbf{\tilde{z}}^i_1,\cdots,\mathbf{\tilde{z}}^i_{n_t}\}$ and $\mathbf{\tilde{Z}}^j=\{\mathbf{\tilde{z}}^j_1,\cdots,\mathbf{\tilde{z}}^j_{n_t}\}$ . The similarity of these two videos is calculated as following:
\begin{align}
	S{(\mathbf{X}_i,\mathbf{X}_j)}=\frac{1}{n_t}\sum_{t=1}^{n_t}\frac{(\mathbf{\tilde{z}}^i_t)^\intercal\mathbf{\tilde{z}}_t^j}{||\mathbf{\tilde{z}}^i_t ||||\mathbf{\tilde{z}}_t^j||}
\end{align}
where we use the cosine distance in computing the similarity of each frame pair. We use $S{(\mathbf{X}_i,\mathbf{X}_j)}$ for the video sequence matching in few-shot classification.

\begin{table*}[thbp]
	\centering
	% \begin{threeparttable}
	\resizebox{0.70\textwidth}{!}{
		\begin{tabular}{l|c|lllll}
			\toprule
			\multirow{2}{*}{\textbf{Method}} & \multirow{2}{*}{\textbf{Data Split}}   & \multicolumn{5}{c}{\textbf{Something-Something-V2}}  \\
			\cmidrule{3-7}
			&            &\textbf{1-shot} & \textbf{2-shot } & \textbf{3-shot  } & \textbf{4-shot  }& \textbf{5-shot  } \\
			\midrule
			RGB$^*$        &\multirow{10}{*}{Small}
					         &   20.0 & 25.3 & 29.2 & 30.9 & 33.6 \\
		    Flow$^*$     &       &   21.2 & 26.0 & 30.1 & 31.8 & 33.8 \\
		    LSTM$^*$      &      &   19.8 & 24.9 & 28.6 & 30.6 & 32.5 \\
		    Nearest-FT$^*$ &       &   27.5 & 32.0 & 35.9 & 37.8 & 41.0 \\
		    Nearest$^*$    &        &   28.1 & 33.3 & 37.2 & 39.2 & 43.8 \\
		    MatchingNet\cite{vinyals2016matching}$^*$ &     &   31.3 & 35.9 & 39.8 & 40.5 & 45.5 \\
%			MatchingNet\cite{vinyals2016matching}$^\dagger$&    & 33.5\colorpm{$\pm$0.3} & 37.0\colorpm{$\pm$0.2} & 40.0\colorpm{$\pm$0.3}& 41.8\colorpm{$\pm$0.2}& 43.0\colorpm{$\pm$0.2}  \\
%			MatchingNet$^\dagger$&    & 33.2\colorpm{$\pm$0.3} &  \\
			MAML\cite{finn2017model}      &      &   30.9 & 35.1 & 38.6 & 40.0 & 41.9 \\
			Plain CMN\cite{zhu2018compound} &       &   33.4 & 38.9 & 42.5 & 44.0 & 46.5 \\
			LSTM-EMB$^*$    &        &   33.0 & 38.5 & 41.8 & 43.8 & 46.2 \\
			CMN\cite{zhu2020label}        &       &   36.2 & 42.1 & 44.6 & 47.0 & 48.8 \\
			\cmidrule{1-1}\cmidrule{3-7}
			\textbf{ITANet} &    &   \textbf{39.8\colorpm{$\pm$0.2}}  & \textbf{45.8\colorpm{$\pm$0.2}}    & \textbf{49.4\colorpm{$\pm$0.4}} &  \textbf{51.8\colorpm{$\pm$0.3}} &
			   \textbf{53.7\colorpm{$\pm$0.2}} \\
			\midrule
			TSN++$^\ddag$         & \multirow{6}{*}{Full}
										       & 33.6 &   -  &     -   &   - & 43.0 \\
			CMN++$^\ddag$        &    & 34.4 &   -  &     -   &   - & 43.8\\
			TRN++$^\ddag$       &    & 38.6 &   -  &     -   &   - & 48.9\\
			OTAM\cite{cao2020few}$^\ddag$      &
					        &  42.8 &   -  &     -   &   - & 52.3\\
%			MatchingNet$^\dagger$
%							&     & 42.8\colorpm{$\pm$0.4}  & - & - & - & 54.6\colorpm{$\pm$0.3}\\
		    \cmidrule{1-1}\cmidrule{3-7}
			\textbf{ITANet}    &   & \textbf{49.2\colorpm{$\pm$0.2}} & \textbf{55.5\colorpm{$\pm$0.3}} &\textbf{59.1\colorpm{$\pm$0.2}}  & \textbf{61.0\colorpm{$\pm$0.3}}  &  \textbf{62.3\colorpm{$\pm$0.3}}  \\
			\bottomrule
	\end{tabular}} %\end{threeparttable}
		\vspace{-0.8em}
		\caption{\textbf{Accuracy on Something-Something-V2 dataset.} We use ResNet-50 as the backbone for fair compassion with previous worsk. $*$ means the results are copied from \protect\cite{zhu2020label} directly. $\ddag$ means the results are copied from \protect\cite{cao2020few}.}
	\label{tab:quant_sth}
\end{table*}
\subsection{Meta Learning with Multi-task Loss}\label{sec:learning}
To estimate the model parameters, we train our network in the meta-leanring framework. Inspired by previous works~\cite{liu2020part,yan2019dual}, we propose a multi-task learning paradigm to train the meta model in order to obtain better video representation.

Specifically, our learning objective consists of two losses: a) the per-task classification loss defined on the episode label space, b) the semantic video classification loss on the global video class space $\mathcal{C}^{tr}$.
For the per-task classification, the final prediction for a query video $\mathbf{X}^q$ in $C$-way-$K$-shot setting is generated by:
\begin{align}
	\tilde{y}^q=\sum_{i=1}^{CK} S{(\mathbf{X}^s_i,\mathbf{X}^q)} y^s_i
\end{align}
where $y^s_i$ is the label of the $i$-th support video.
For the semantic video classification, we add a linear classification head after the context encoding network to predict the global label $\tilde{y}_{sem}$ for all videos in each task.
 
We adopt the standard cross-entropy loss in both loss terms for training our model on the meta-training set $\mathcal{D}^{tr}$, and the loss for each task can be written as:
\begin{align}
	\mathcal{L}_{all}=\mathcal{L}_{meta}+\beta\mathcal{L}_{sem}
	\end{align}
where $\mathcal{L}_{sem}=\mathcal{L}_{CE}(\tilde{y}^s_{sem},y^s_{sem})+\mathcal{L}_{CE}(\tilde{y}_{sem}^q,y_{sem}^q)$ and $\mathcal{L}_{meta}=\mathcal{L}_{CE}(\tilde{y}^q,y^q)$. Here $\beta$ is the hyper-parameters to balance the ratio of the meta loss and semantic loss. To compare different learning paradigms, we perform extensive experiments and detailed analysis in Sec~\ref{sec:ablation}.

\section{Experiments}\label{sec:cls_experiment}
%To evaluate the performance of our proposed model for few-shot video classification
%We conduct extensive experiments on two datasets following previous works. 
In this section, we conduct a series of experiments to validate the effectiveness of our method.
Below we first give a brief introduction of experimental configurations and report the quantitative results on two benchmarks in Sec.~\ref{sec:configuration}. Then we conduct ablative experiments to show the efficacy of our model design in Sec.~\ref{sec:ablation}.

\subsection{Results on SthSth-V2 and Kinetics}\label{sec:configuration}
\paragraph{Datasets}
Following previous works, we use the Kinetics~\cite{Carreira_2017_CVPR} and Something-Something V2~\cite{goyal2017something} as the benchmarks. For the Kinetics dataset, we follow the same split as CMN~\cite{zhu2018compound}, which samples 64 classes for meta training, 12 classes for validation, and 24 classes for meta testing. Each class consists of round 100 samples. 

For the Something-Something V2 dataset, there exist two splits proposed by ~\cite{cao2020few} and ~\cite{zhu2020label}, both consisting of 64, 12, and 24 classes as the meta-training, meta-validation, and meta-testing set. The two splits differ in the number of samples for each class. There are about 100 samples per class in the ~\cite{zhu2020label}'s split, denoted as "SthV2-\textit{small}" in the following paragraphs. And ~\cite{cao2020few}'s split uses all annotated videos of the original Something-Something V2, which is denoted as "SthV2-\textit{full}" below.

%As pointed out by ~\cite{zhou2018temporal} ~\cite{xie2018rethinking}, existing action recognition datasets can be roughly classified into two groups: YouTube type videos: UCF101 ~\cite{soomro2012ucf101}, Sports 1M ~\cite{karpathy2014large}, Kinetics ~\cite{kay2017kinetics}, and crowd-sourced videos: Charades ~\cite{sigurdsson2016hollywood}, Something-Something V1\&V2 ~\cite{goyal2017something}, in which the videos are collected by asking the crowd-source workers to record themselves performing instructed activities. Crowdsourced videos usually focus more on modeling the temporal relationships, since visual contents among different classes are more similar than those of YouTube type videos. Following previous work ~\cite{cao2020few} ~\cite{zhu2020label} , we evaluate our model on two action recognition datasets, Kinetics ~\cite{kay2017kinetics} and Something-Something V2 ~\cite{goyal2017something}.

\paragraph{Experimental Configuration}

%For a $C$-way, $K$-shot test setting, we randomly sample $C$ classes with each class containing $k$ examples as the support set.
%We construct the query set to have 1 examples, where each unlabeled sample in the query set belongs to one of the $n$ classes in the support set. 
%Thus each episode has a total of $n(k + 1)$ examples. 
%We conduct 5-way-1-shot and 
%For network details, we adopt ResNet, initialized with weights pre-trained on ImageNet, as feature extractor to compute the convolutional feature maps. 
We follow the same video preprocessing procedure as OTAM~\cite{cao2020few}. During training, we first resize each frame in the video to 256$\times$256 and then randomly crop a 224$\times$224 region from the video clip. We sparsely and uniformly sample $n_t=8$ frames per video. For inference, we change the random crop to center crop. For the Kinetics dataset, horizontal flip is applied as data augmentation during training. For the Something-Something V2 dataset, as pointed out in ~\cite{cao2020few}, the dataset is sensitive to concepts of left and right, hence we do not use horizontal flip for this dataset.

Following the experiment settings and learning schedule from ~\cite{zhu2018compound} ~\cite{cao2020few}, we use ResNet-50 ~\cite{he2016deep} as the backbone network which is initialized with pre-trained weights on ImageNet ~\cite{deng2009imagenet}. We optimize our model with SGD. We use the meta-validation set to tune the parameters.
% and stop the training process when the accuracy of the meta-validation set is about to decrease. 
 We perform different $C$-way $K$-shot experiments on the two datasets, with 95\% confidence interval in the meta-test phase. 
Specifically, the final results are reported over 5 runs and we randomly sample 20,000 episodes for each run. %More detailed information can be found in the supplementary.
%with a starting learning rate of 0.001 and decaying every 30 epochs by 0.1 We implemented our framework with PyTorch [31]. The full model trains for 10 hours on 4 TITAN Xp GPUs.

%%%%%%%%%%%%%%%%%%%%%%%%%%%%%%%%%%%%%%%%%%%%%%%%%%%%%%%%%%%%%%%%%%%%%%%%%%%%%%%%%%%%%
%  Quantity_kin
%%%%%%%%%%%%%%%%%%%%%%%%%%%%%%%%%%%%%%%%%%%%%%%%%%%%%%%%%%%%%%%%%%%%%%%%%%%%%%%%%%%%%

%%%%%%%%%%%%%%%%%%%%%%%%%%%%%%%%%%%%%%%%%%%%%%%%%%%%%%%%%%%%%%%%%%%%%%%%%%%%%%%%%%%%%

% TODO: list the backbone, pretrained model, and where the results come from!
% We compare our method with xxx,xxx,xxx.
\paragraph{Quantitative Comparison}
We compare our algorithm with a wide range of recent methods, as shown in Tab~\ref{tab:quant_sth} and Tab~\ref{tab:quant_kin},
%including OTAM, TSN++, CMN++, TRN++ reported by~\cite{cao2020few}, CMN, LSTM-EMB, MAML, MatchingNet reported by ~\cite{zhu2020label} and ARN ~\cite{zhang2020few} .
and report the 5-way performance on 2 benchmarks, including 3 splits: SthV2-\textit{small}, SthV2-\textit{full} and Kinetics-100. 
%Our method achieves the state of-the-art on all benchmarks. 

 The actions in SthSth-v2 are typically order-sensitive(\textit{e.g.} \textit{pushing something from left to right} and \textit{pushing something from right to left}), and more challenging to recognize, which requires more accurate and robust temporal alignment for classification. We report the performance of small/full split in Tab~\ref{tab:quant_sth}. We achieve $\mathbf{39.8\%}/\mathbf{53.7\%}$ for the 5-way 1/5-shot setting on SthV2-\textit{small}, which outperform the state-of-the-art methods by a sizable margin of $\mathbf{3.6}$ and $\mathbf{4.9}$, respectively. On the full split, our method benefits from more labeled samples in meta-train phase, and achieve $\mathbf{49.2\%}/\mathbf{62.3\%}$ for 5-way 1/5-shot in meta-test phase. The significant improvements demonstrate the effectiveness of our proposed method.
 
 Further, we evaluate our method on the Kinetics dataset, in which the videos are not order-sensitive, and report the performance in Tab~\ref{tab:quant_kin}. The results show our method is also competitive compared with recent works.
 
%Even compared with the latest works, the improvement is still significant, which indicates the learned knowledge of implicit alignment is generalizable for novel classes.
\begin{table}[tb!]
	\centering
	% \begin{threeparttable}
	\resizebox{0.43\textwidth}{!}{
		\begin{tabular}{l|c|ccccc}
			\toprule
			\multirow{1}{*}{\textbf{Method}}  &\textbf{5-Way 1-Shot} & \textbf{5-Way 5-Shot} \\
			\midrule
			%RGB\cite{zhu2020label}         & 28.7   & 48.6     \\
		    %Flow\cite{zhu2020label}           & 24.4   & 33.1   \\
		    %LSTM\cite{zhu2020label}           & 28.9   & 49.0    \\
		    %Nearest-FT\cite{zhu2020label}     & 48.2   & 62.6    \\
		    %Nearest\cite{zhu2020label}        & 51.1   & 68.9    \\
		    MatchingNet\cite{vinyals2016matching}    & 53.3   & 74.6    \\
%		    			MatchingNet$^\dagger$ \\
			MAML\cite{finn2017model}          & 54.2   & 75.3     \\
			%Plain CMN\cite{zhu2018compound}       & 57.3   & 76.0     \\
			LSTM-EMB\cite{zhu2020label}        & 57.6   & 76.2     \\
			CMN\cite{zhu2020label}            & 60.5   & 78.9    \\
			ARN\cite{zhang2020few}            & 63.7   & 82.4    \\
			Mean\cite{cao2020few}$\dagger$             & 67.8   & 78.9    \\
			Plain DTW\cite{cao2020few}$\dagger$           & 69.2   & 80.6 \\
			OTAM\cite{cao2020few}$\dagger$              & 73.0   & \textbf{85.8}    \\
%			OTAM$^\ddag$             & 72.0\colorpm{$\pm$ 0.3}  &  84.5\colorpm{$\pm$ 0.2}    \\
			\midrule
			\textbf{ITANet}					&\textbf{73.6\colorpm{$\pm$ 0.2}} & 84.3\colorpm{$\pm$ 0.3}\\
			\bottomrule
	\end{tabular}} %\end{threeparttable}
		\vspace{-0.8em}
		\caption{\textbf{Results on Kinetics-100 dataset.} $\dagger$ means the results are copied from \protect\cite{cao2020few}.
%		$\ddag$ is our re-implement OTAM.
		}
	\label{tab:quant_kin}
\end{table}
% and $0$ on SthV2-Full, which outperform the state-of-the-art methods by a significant margin of $0$ and $0$, respectively.

%The results can be found in Tab~\ref{tab:quant_sth} and Tab~\ref{tab:quant_kin}.

% TODO: list the backbone, pretrained model, and where the results come from!
% We compare our method with xxx,xxx,xxx.

\subsection{Ablation Study}
We conduct several ablative experiments to evaluate the the effectiveness of our model on the SthV2-\textit{small}.
%\paragraph{Effectiveness of Model Components}

\paragraph{Analysis of Model Components}\label{sec:ablation}
 % the Implicit Alignment
% 1.temperature(?) 2. ?(channel attention? spatial attention? 3. implicit alignment 4.multitask
We first show the importance of each component used in our model by a series of ablation studies.
%Our model design for the few-shot video classification task consists of a context encoding module followed with an implicit temporal alignment module. 
Specifically, we show the effectiveness of each component under the 5-way 1-shot setting in Tab~\ref{tab:model_ablation}. We can see that the spatial and channel context can improve the baseline method with an improvement of $\mathbf{0.5\%}$ and $\mathbf{1.2\%}$, respectively, which indicate the frame-level context and video-level context make the representation more discriminative. The proposed implicit temporal alignment contributes to a significant improvement of $\mathbf{5.4\%}$ over the baseline by exploring the temporal order information effectively. The multi-task learning paradigms introduced for few-shot video learning is able to improve the final performance with a sizable margin at $\mathbf{1.9\%}$, which demonstrates the effectiveness of exploiting semantic information in our method. All components collaborate and complement each other, reaching $\mathbf{39.8\%}$ finally.
%exploits the  efficiently. As a reminder for the benifit of future research, here we also reveal the boost of scaling the cosine score, which the prior works have probably neglected.

%\begin{figure}[t]
%    \centering
%    \includegraphics[width=0.8\linewidth]{example-image-a}
%%    \vspace{-2em}
%    \caption{Visualization analysis.}\vspace{-3mm}
%    \label{fig:overview}
%\end{figure}	
\paragraph{The Effect of Learning Paradigms} % Different Learning Paradigms
To study the impact of different learning strategies, we conduct several experiments on a MatchingNet baseline and report the results in Tab~\ref{tab:multitask_ablation}.
%First, we only use the semantic loss to train the embedding network based on the global semantic label space for all videos in on train split, then the linear layer is ignored and the videos are embedded into a global feature vector for Matching Network.
%Second, we follow the standard meta learning protocol and use the meta loss to train the baseline method(without context modeling and implicit alignment). 
First, we only use the semantic loss to train the embedding network or the standard meta loss and learning protocol to train the baseline model. 
Then, we explore learning with semantic loss in the first stage, followed by learning with meta loss in the second stage. Finally, multi-task learning strategy are used to verify the effective of the semantics information used in a joint learning manner. From the experimental results, we find the multi-task learning achieve the best results among the different learning strategy, which indicates it can exploit the global semantic label effectively and benefit from the joint optimization.

% Similar ideas are claimed effective in some recent few-shot learning works ~\cite{chen2019closer}. Second, \textbf{\textit{Episodic-only}}: the standard episodic learning paradigm as described in ~\cite{vinyals2016matching}. Third, \textbf{\textit{Semantic+Episodic}}: sequentially perform \textit{Semantic-only} and then \textit{Episodic-only}

%Third, \textbf{\textit{Joint}}: jointly optimize the semantic-loss and meta-loss, which can be seen as a combination of the first 2 paradigms. We empirically find it helpful to perform multitask learning, i.e. to add semantic loss during meta-training for the few-shot video classification tasks. The results are reported in Tab~\ref{tab:multitask_ablation}.
	% \textbf{Semantic-Cosine}: Instead of 

\paragraph{Design of Implicit Temporal Alignment}\label{sec:ablation}
Here we show the effectiveness of each component of implicit temporal alignment by adding them progressively onto our baseline model. From the Tab~\ref{tab:implicit_align_ablation}, we find the temporal relation improves the baseline from 33.2 to 35.0, which indicates the temporal contextual relation helps the similarity measurement. Moreover, adding position encoding enables the implicit alignment process temporal order-aware, and has a gain of 0.9. Finally, we use the frame-wise similarity to reduce the noise in the similarity of global averaged feature, which achieves the further improvement and reach $\mathbf{38.6\%}$.
%\paragraph{Scaling the Cosine Score}  % Empirical study of
%show reimplemented results.

%%%%%%%%%%%%%%%%%%%%%%%%%%%%%%%%%%%%%%%%%%%%%%%%%%%%%%%%%%%%%%%%%%%%%%%%%%%%%%%%%%%%%
%  Model ablation
%%%%%%%%%%%%%%%%%%%%%%%%%%%%%%%%%%%%%%%%%%%%%%%%%%%%%%%%%%%%%%%%%%%%%%%%%%%%%%%%%%%%%
\begin{table}[t!]
	\centering
	\resizebox{0.42\textwidth}{!}{
		\begin{tabular}{cccc|ccc}
			\toprule
%\textbf{Spatial} & \textbf{Channel} & \textbf{ImAlign}  & \textbf{MultiTask}    &  \textbf{C5-K1}                   & \textbf{C5-K5}          \\
%				\midrule
%	   &    &      &    &      33.2\colorpm{$\pm$0.3}     &   00.0\colorpm{$\pm$0.2}  \\
%\midrule
%\cmark &    &     &   &   33.7\colorpm{$\pm$0.2}   &  -  \\
%	   & \cmark &     &   &   34.4\colorpm{$\pm$0.2}  &  -  \\
%	   &  & \cmark   &   &     38.6\colorpm{$\pm$0.3}   &   - \\
%	   &  & & \cmark   &     35.1\colorpm{$\pm$0.3}   &   - \\
%\midrule
%\cmark & \cmark    & \cmark   & \cmark &   39.8\colorpm{$\pm$0.2}     &  -  \\
\textbf{Spatial} & \textbf{Channel} & \textbf{ImAlign}  & \textbf{MultiTask}    &  \textbf{5-Way 1-Shot}                     \\
				\midrule
	   &    &      &    &      33.2\colorpm{$\pm$0.3}      \\
\midrule
\cmark &    &     &   &   33.7\colorpm{$\pm$0.2}    \\
	   & \cmark &     &   &   34.4\colorpm{$\pm$0.2}    \\
	   &  & \cmark   &   &     38.6\colorpm{$\pm$0.3}    \\
	   &  & & \cmark   &     35.1\colorpm{$\pm$0.3}     \\
\midrule
\cmark & \cmark    & \cmark   & \cmark &   \textbf{39.8\colorpm{$\pm$0.2}}      \\
\bottomrule
		\end{tabular}
	}
	    \vspace{-0.8em}
		\caption{\textbf{Ablation of the Model Components on SthV2-\textit{small}.} We use $\text{MatchingNet}$ as the baseline(our re-implemented matching network without FCE), which is exactly the first row.}\vspace{-1mm}
	\label{tab:model_ablation}
\end{table}
%%%%%%%%%%%%%%%%%%%%%%%%%%%%%%%%%%%%%%%%%%%%%%%%%%%%%%%%%%%%%%%%%%%%%%%%%%%%%%%%%%%%%
%  Multitask ablation
%%%%%%%%%%%%%%%%%%%%%%%%%%%%%%%%%%%%%%%%%%%%%%%%%%%%%%%%%%%%%%%%%%%%%%%%%%%%%%%%%%%%%
%\begin{table}[t!]
%	\centering
%	\resizebox{0.45\textwidth}{!}{
%		\begin{tabular}{c|c|c|c|ccc}
%			\toprule
%				 \textbf{Standard} & \textbf{Episodic}  & \textbf{Joint}    &  \textbf{N5-K1}                   & \textbf{N5-K5}          \\
%				\midrule
%				\cmark & \xmark    & \xmark &   69.8         &    \\
%				\xmark & \cmark    & \xmark &     71.5       &    \\
%				\cmark & \cmark    & \xmark &       71.4     &    \\
%				\midrule
%				\cmark & \cmark    & \cmark &     72.0       &    \\
%%				\midrule
%%				\textbf{DisAlgin}                &  61.7\colorpm{$\pm0.00$}  &  65.0\colorpm{$\pm0.00$}    & 65.8\colorpm{$\pm0.00$}    & 69.0\colorpm{$\pm0.00$} \\
%%				\textbf{DisAlgin$^\dagger$}                &  61.7\colorpm{$\pm0.00$}  &  65.0\colorpm{$\pm0.00$}    & 65.8\colorpm{$\pm0.00$}    & 69.0\colorpm{$\pm0.00$} \\
%			\bottomrule
%		\end{tabular}
%	}
%	    \vspace{-0.8em}
%		\caption{\textbf{Results of different learning paradigms on SthSth-V2-small.} We use our dot-product as the baseline in these experiments. 'N5-K1' means 5-way 1-shot setting. }\vspace{-3mm}
%	\label{tab:multitask_ablation}
%\end{table}

\begin{table}[t!]
	\centering
	\resizebox{0.32\textwidth}{!}{
		\begin{tabular}{c|c|c }  %{c|c|c|c|ccc}
			\toprule
%				 \textbf{Method}    \\
			\multicolumn{2}{c}{\textbf{Learning Method}} &  \multirow{2}{*}{\textbf{ 5-Way 1-Shot}  }                         \\
			\cmidrule{1-2}
			  \textbf{Stage-1} &  \textbf{Stage-2}           & \\
				\midrule
				Meta Loss  & \xmark    &   33.2\colorpm{$\pm$0.3}       \\
				Semantic Loss& \xmark  &     32.2\colorpm{$\pm$0.2}    \\
				Semantic Loss & Meta Loss  & 34.4\colorpm{$\pm$0.2}         \\
				\midrule
				Multi Task & \xmark     &   \textbf{35.1\colorpm{$\pm$0.3}}      \\
%				\midrule
%				\textbf{DisAlgin}                &  61.7\colorpm{$\pm0.00$}  &  65.0\colorpm{$\pm0.00$}    & 65.8\colorpm{$\pm0.00$}    & 69.0\colorpm{$\pm0.00$} \\
%				\textbf{DisAlgin$^\dagger$}                &  61.7\colorpm{$\pm0.00$}  &  65.0\colorpm{$\pm0.00$}    & 65.8\colorpm{$\pm0.00$}    & 69.0\colorpm{$\pm0.00$} \\
			\bottomrule
		\end{tabular}
	}
	    \vspace{-0.8em}
		\caption{\textbf{Different Learning paradigms on SthV2-\textit{small}.} }\vspace{-1mm}
	\label{tab:multitask_ablation}
\end{table}

%%%%%%%%%%%%%%%%%%%%%%%%%%%%%%%%%%%%%%%%%%%%%%%%%%%%%%%%%%%%%%%%%%%%%%%%%%%%%%%%%%%%%
%  Model ablation
%%%%%%%%%%%%%%%%%%%%%%%%%%%%%%%%%%%%%%%%%%%%%%%%%%%%%%%%%%%%%%%%%%%%%%%%%%%%%%%%%%%%%
\begin{table}[t!]
	\centering
	\resizebox{0.41\textwidth}{!}{
		\begin{tabular}{ccc|cc}
			\toprule
				 \textbf{TempRelation}  & \textbf{PosEncoding}  & \textbf{Frame-wise}    &  \textbf{5-Way 1-Shot}                            \\
				\midrule
				       &         &        &     33.2\colorpm{$\pm$0.3}   \\
				\midrule
				\cmark  &        &          &      35.0\colorpm{$\pm$0.2}    \\
				\cmark  & \cmark &            &    35.9\colorpm{$\pm$0.3}   \\
				\cmark  & \cmark & \cmark       &     \textbf{38.6\colorpm{$\pm$0.3}}    \\
%				\midrule
%				\cmark & \cmark    & \cmark   &   -     &  -  \\
%				\midrule
%				\textbf{DisAlgin}                &  61.7\colorpm{$\pm0.00$}  &  65.0\colorpm{$\pm0.00$}    & 65.8\colorpm{$\pm0.00$}    & 69.0\colorpm{$\pm0.00$} \\
%				\textbf{DisAlgin$^\dagger$}                &  61.7\colorpm{$\pm0.00$}  &  65.0\colorpm{$\pm0.00$}    & 65.8\colorpm{$\pm0.00$}    & 69.0\colorpm{$\pm0.00$} \\
			\bottomrule
		\end{tabular}
	}
	    \vspace{-0.8em}
		\caption{\textbf{Ablation of the Implicit Temporal Alignment on SthV2-\textit{small}.} We use $\text{MatchingNet}$ as the baseline.}\vspace{-2mm}
	\label{tab:implicit_align_ablation}
\end{table}

\section{Conclusion}\label{sec:conclusion}

In this paper, we have presented a novel matching-based method for few-shot video classification.  
%aims to learn to classify new action videos with only a few labeled examples, which has a wide range of realistic applications. Previous methods either focus on improving feature representation with sophisticated model or metric learning with heuristic design. In this work, we propose a novel few-shot video action recognition framework based on the matching-based method. 
To cope with large intra-class variation in videos, we developed a novel video representation that augments conv features of videos with spatial, feature channel and temporal context based on a factorized self-attention mechanism. This enables us to achieve an implicit temporal alignment in video feature space and adopt a simple frame-wise similarity metric for video matching. We also study a multi-task learning strategy and demonstrate its superiority in few-shot video learning. Our method achieves competitive results on the Kinetics and outperforms the previous works with a sizable margin on the more challenging Something-Something-V2 benchmark.
 
 \newpage
\small
\bibliographystyle{named}
\bibliography{ijcai21}
\appendix
\newpage
\section{Cross Dataset Experiments}

We conduct the experiments to investigate the transfer capacity by evaluating the model trained on Kinetics-100 on SthV2 and vice versa. The performance is reported in Tab.~\ref{tab:transfer}. From the table, we find that our method can also achieve significantly improvements on this transfer learning setting, which demonstrates the superiority of our model. We will add detailed discussion in the revision.

\begin{table}[h!]
	\centering
	%	\vspace{-.15em}
	\resizebox{0.85\linewidth}{!}{
		\begin{tabular}{c|c|c|c|c}
			\toprule
			\textbf{Method} & \textbf{Source} & \textbf{Target} & \textbf{5W-1S} & \textbf{5W-5S} \\
			\midrule
			\multirow{2}{*}{Baseline} & Kinetics  & SthV2-Small & 29.8\colorpm{$\pm0.4$}&38.4\colorpm{$\pm 0.2$}  \\
			% & Kinetics   &  SthV2-Large   &    \\
			\cmidrule{2-5}
			& SthV2-Small  &  Kinetics & 40.7\colorpm{$\pm 0.3$} & 53.5\colorpm{$\pm0.3$} \\
			% \cmidrule{2-5}
			% &  SthV2-Large & Kinetics  &   \\
			\midrule
			\multirow{2}{*}{\textbf{ITANet}} & Kinetics  & SthV2-Small &  30.9\colorpm{$\pm 0.1$} & 40.1\colorpm{$\pm 0.5$} \\
			% & Kinetics   &  SthV2-Large   &     \\
			\cmidrule{2-5}
			& SthV2-Small  &  Kinetics &  56.8\colorpm{$\pm 0.1$}  & 70.9\colorpm{$\pm 0.2$}\\
			% \cmidrule{2-5}
			% &  SthV2-Large & Kinetics  &   \\
			\bottomrule
	\end{tabular}}
	\caption{\textbf{Experiments of Cross Datasets Generalization.} We adopt the Matching Network as the baseline. We report the top-1 accuracy of 5-way 1-shot and 5-way 5-shot setting.\vspace{-1em}}
	\label{tab:transfer}
	%\end{center}
\end{table}

\section{Ablative Study on Number of Frames}
% \textbf{Reviewer \#60}
First, we choose $t=8$ as the default setting for a fair comparison with previous works. To further investigate the influence of the number of sampled frames, we test three different settings as shown in the table below. We find increasing the number of frames will consistently improve the accuracy, while this also introduces larger matching cost.
\begin{table}[h]
	\centering
%	\vspace{-.15em}
	\resizebox{0.75\linewidth}{!}{
		\begin{tabular}{c|c|c|c}
			\toprule
			\textbf{Method} & \textbf{Frame Number}  & \textbf{5W-1S} & \textbf{5W-5S} \\
			\midrule
			\multirow{3}{*}{ITANet} & 4  & 46.3\colorpm{$\pm 0.2$} & 58.4\colorpm{$\pm 0.1$}    \\
			\cmidrule{2-4}
% 			\multirow{2}{*}{ITANet} 
			  & 8   & 49.2\colorpm{$\pm0.2$}  & 62.3\colorpm{$\pm0.3$}     \\
			  \cmidrule{2-4}
			  & 16 &  49.5\colorpm{$\pm0.1$}  & 64.3\colorpm{$\pm 0.2$}  \\
			 % \cmidrule{2-5}
			 % &  SthV2-Large & Kinetics  &   \\
			  \bottomrule
	\end{tabular}}
	
	\caption{\textbf{Ablation Study on Frame Number.} We conduct all experiments on SthV2-large dataset.\vspace{-1em}}
    \label{tab:num_frames}
	%\end{center}
\end{table}

\section{Matching Efficiency} 
We would like to clarify that although our feature enhancement indeed introduces additional cost, our ITANet achieves higher efficiency in the video matching stage. Specifically, in OTAM, each query video requires a dense matching with support videos with a quadratic cost with respect to $T$. Our solution models temporal relation by utilizing self-attention for each video only once, followed by a distance calculation whose cost is linear with respect to $T$. %the cost of calculation distance between two videos is linear on $T$.
We line up the computation complexity of the temporal alignment for a clear comparison.
\begin{table}[h!]
	\centering
%	\vspace{-.15em}
	\resizebox{0.8\linewidth}{!}{
		\begin{tabular}{c|l}
			\toprule
			\textbf{Method} & \textbf{Complexity}  \\
			\midrule
			OTAM & $\mathcal{O}(NK*Q*C*T^2)$ \\
			\midrule
			\textbf{ITANet}  & $\mathcal{O}((NK+Q)*C*T^2 +NK*Q*C*T)$      \\
			 % \cmidrule{2-5}
			 % & SthV2-Small  &  Kinetics &  \\
			 % \cmidrule{2-5}
			 % &  SthV2-Large & Kinetics  &   \\
			\bottomrule
	\end{tabular}}
	\caption{\textbf{Complexity of Temporal Alignment.} 'N' and 'K': number of way and shot in support set, 'Q': number of query videos, 'T': number of frames of each video, 'C': feature dimension.	\vspace{-1em}}
    \label{tab:complexity}
	%\end{center}
\end{table}
%Especially, our implicit align strategy is superior when then support has many way(N is large) or query set has a large number of videos(Q is large), which indicates that ITANet has better scalability as the growth of test samples or number of classes.

We note that the temporal self-attention in ITANet is a one-time cost for each support/query video, while OTAM requires costly comparison for each support-query pair, indicating that ITANet has better scalability as the growth of test samples or number of classes (i.e., when Q or N is large).
\end{document}